\documentclass[11pt,a4paper]{article}
\usepackage[hyperref]{emnlp2020}
\usepackage{times}
\usepackage{latexsym}
\usepackage{CJKutf8}

\usepackage{url}
\usepackage{graphicx}  
\usepackage{mathrsfs}
\usepackage{amsmath}
\usepackage{bm}
\usepackage{multirow}
\usepackage{subfigure}
\usepackage{makecell}
\usepackage{array}
\usepackage{booktabs}
\usepackage{algorithm}
\usepackage{algpseudocode}

\usepackage{microtype}

\aclfinalcopy 

\title{Transformer-GCRF: Recovering Chinese Dropped Pronouns with General Conditional Random Fields}

\author{Jingxuan Yang$^1$, Kerui Xu$^1$, Jun Xu$^{2,3*}$, Si Li$^{1*}$, Sheng Gao$^1$, Jun Guo$^1$, \\
\textbf{Ji-Rong Wen$^{2,3}$ and Nianwen Xue$^4$} \\
$^1$School of Artificial Intelligence, Beijing University of Posts and Telecommunications \\
$^2$Gaoling School of Artificial Intelligence, Renmin University of China\\
$^3$Beijing Key Laboratory of Big Data Management and Analysis Methods\\
$^4$Department of Computer Science, Brandeis University \\
{\tt \{yjx, xukerui, lisi, gaosheng, guojun\}@bupt.edu.cn} \\
{\tt junxu@ruc.edu.cn, jirong.wen@gmail.com, xuen@brandeis.edu} 
  }

\date{}

\begin{document}
\begin{CJK*}{UTF8}{gbsn}
\maketitle
\begin{abstract}
Pronouns are often dropped in Chinese conversations and recovering the dropped pronouns is important for NLP applications such as Machine Translation. Existing approaches usually formulate this as a sequence labeling task of predicting whether there is a dropped pronoun before each token and its type. Each utterance is considered to be a sequence and labeled independently. Although these approaches have shown promise, labeling each utterance independently ignores the dependencies between pronouns in neighboring utterances. Modeling these dependencies is critical to improving the performance of dropped pronoun recovery. In this paper, we present a novel framework that combines the strength of Transformer network with General Conditional Random Fields (GCRF) to model the dependencies between pronouns in neighboring utterances. Results on three Chinese conversation datasets show that the Transformer-GCRF model outperforms the state-of-the-art dropped pronoun recovery models. Exploratory analysis also demonstrates that the GCRF did help to capture the dependencies between pronouns in neighboring utterances, thus contributes to the performance improvements.\let\thefootnote\relax\footnotetext{$^*$ Corresponding author}
\end{abstract}

\section{Introduction}
In pro-drop languages such as Chinese, pronouns can be dropped as the identity of the pronoun can be inferred from the context, and this happens more frequently in conversations~\citep{Yang:15}. Recovering dropped pronouns (DPs) is a critical task for many NLP applications such as Machine Translation where the dropped pronouns need to be translated explicitly in the target language~\citep{wang:2016a, wang:2016d, wang:2018}. 
Recovering dropped pronoun is different from traditional pronoun resolution tasks~\citep{Zhao:07,Yin:17,yin:2018}, which aim to resolve the anaphoric pronouns to their antecedents. In dropped pronoun recovery, we consider both anaphoric and non-anaphoric pronouns, and we do not directly resolve the dropped pronoun to its antecedent, which is infeasible for non-anaphoric pronouns. We recover the dropped pronoun as one of 17 types pronouns pre-defined in~\citep{Yang:15}, which include five types of abstract pronouns corresponding to the non-anaphoric pronouns. 
Thus traditional rule-based pronoun resolution methods are not suitable for recovering dropped pronouns. 

\begin{figure}
	\centering
	\includegraphics[width=7.6cm, height=5.1cm]{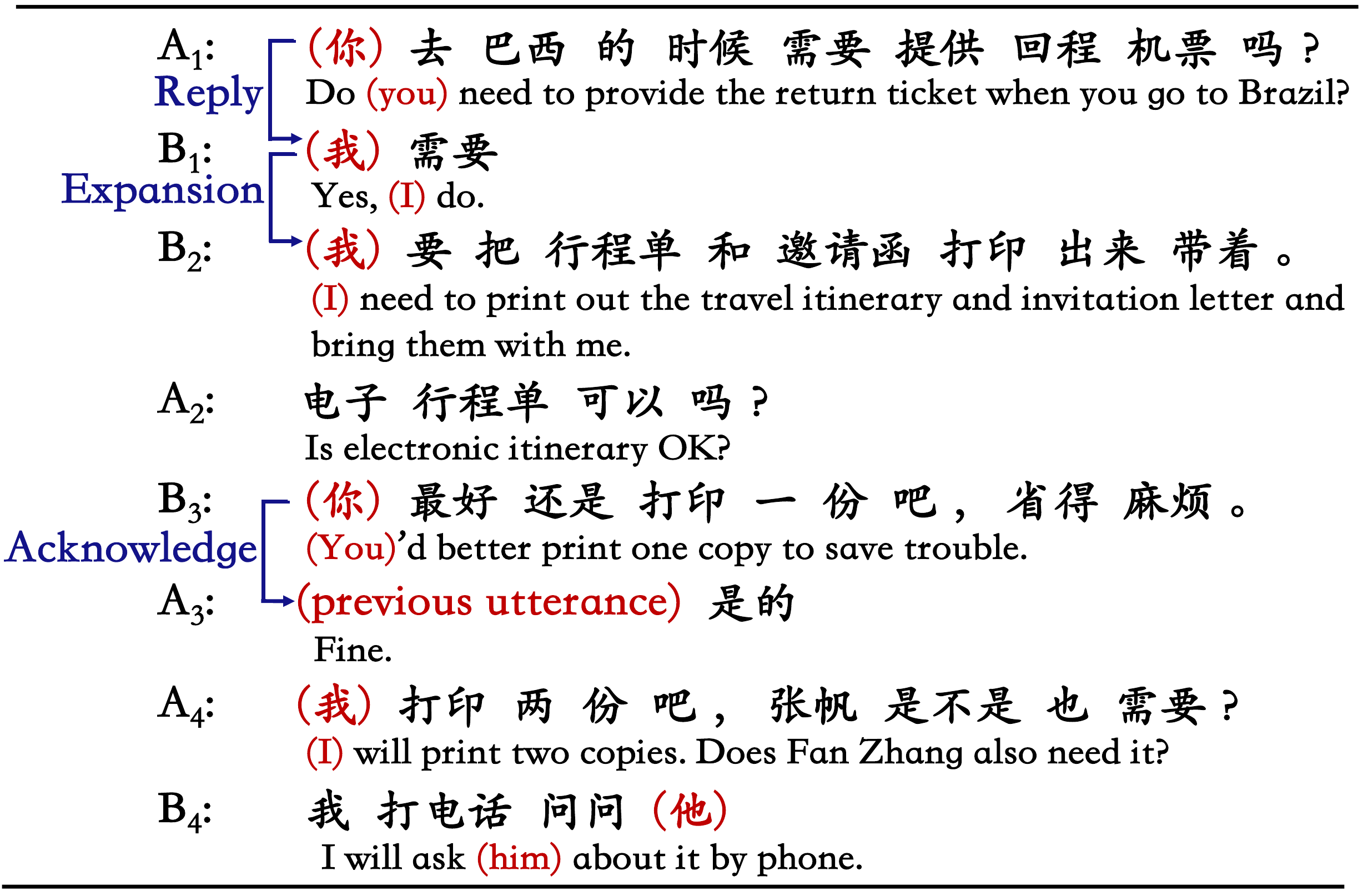}
	\caption{A conversation snippet between participant A and B. The dropped pronouns are shown in the brackets, 
	and the dialogue patterns are marked with blue arrows.
	}
	\label{mode_example}
\end{figure}

Existing approaches formulate  dropped pronoun recovery as a sequence labeling task of predicting whether a pronoun has been dropped before each token and the type of the dropped pronoun. For example,~\citet{Yang:15} first studied this problem in 
SMS data and utilized a Maximum Entropy classifier to recover dropped pronouns. Deep neural networks such as Multi-Layer Perceptrons (MLPs) and structured attention networks have also been used to tackle this problem~\citep{zhang:neural, Yang2019NDPR}. \citet{giannella:17} used a linear-chain CRF to model the dependency between the sequence of predictions in a utterance.

Although these models have achieved various degrees of success, they all assume that each utterance in a conversation should be labeled independently. This practice overlooks the dependencies between dropped pronouns in neighboring utterances, and results in sequences of predicted dropped pronouns are incompatible with one another. We illustrate this problem through an example in Figure~\ref{mode_example}, in which the dropped pronouns are shown in brackets. The pronoun can be dropped as a subject at the beginning of a utterance, or as an object in the middle of a utterance. Pronouns dropped at the beginning of consecutive utterances usually have strong dependencies that pattern with three types of  dialogue transitions (i.e., Reply, Expansion and Acknowledgment) presented in~\citep{xue2016annotating}. For example, in Figure \ref{mode_example}, the pronoun in the second utterance $\mathrm{B}_1$ is ``我~(I)'', the dropped pronoun in the third utterance $\mathrm{B}_2$ should also be ``我~(I)'' since $\mathrm{B}_2$ is an \textit{expansion} of $\mathrm{B}_1$ by the same speaker. Thus modeling the dependency between  pronouns in adjacent sentences is helpful to recover pronoun dropped at utterance-initial positions. In contrast, the pronoun ``他~(him)'' dropped as an object in utterance $\mathrm{B}_4$ should be recovered by capturing referent semantics from the context and modeling token dependencies in the same utterance.

To model the dependencies between predictions in the conversation snippet, we propose a novel framework called Transformer-GCRF that combines the strength of the Transformer model~\cite{vaswani2017attention} in representation learning and the capacity of general Conditional Random Fields (GCRF) to model the dependencies between predictions. 
In the GCRF, a vertical chain is designed to capture the pronoun dependencies between the neighboring utterances, and horizontal chains are used for modeling the prediction dependencies inside each utterance. In this way, Transformer-GCRF successfully models the cross-utterance pronoun dependencies as well as the intra-utterance prediction dependencies simultaneously. Experimental results on three conversation datasets show that Transformer-GCRF significantly outperforms the state-of-the-art recovery models. We also conduct ablative experiments that demonstrate the improvement in performance of our Transformer-GCRF model derives both from the Transformer encoder and the ability of GCRF layer to model the dependencies between dropped pronouns in neighboring utterances. All code is available at \texttt{\url{https://github.com/ningningyang/Transformer-GCRF}}.

The major contributions of the paper are summarized as follows:
\begin{itemize}
\setlength{\itemsep}{-2pt}%
\item We conduct statistical study on pronouns dropped at the beginning of consecutive utterances in conversational corpus, and observe that modeling the dependencies between pronouns in neighboring utterances is important to improve the performance of dropped pronoun recovery. 
\item We propose a novel Transformer-GCRF approach to model both intra-utterance dependencies between predictions in a utterance and cross-utterance dependencies between dropped pronouns in neighboring utterance. The model jointly predicts all dropped pronouns in an entire conversation snippet. 
\item We apply the Transformer-GCRF model on three conversation datasets. Results show that our Transformer-GCRF outperforms the baseline models on all datasets. Exploratory experiments also show that the improvement is attributed to the capacity of the model to capture cross-utterance dependencies.
\end{itemize}

\begin{figure*}
	\centering
	\includegraphics[width=16cm, height=6.3cm]{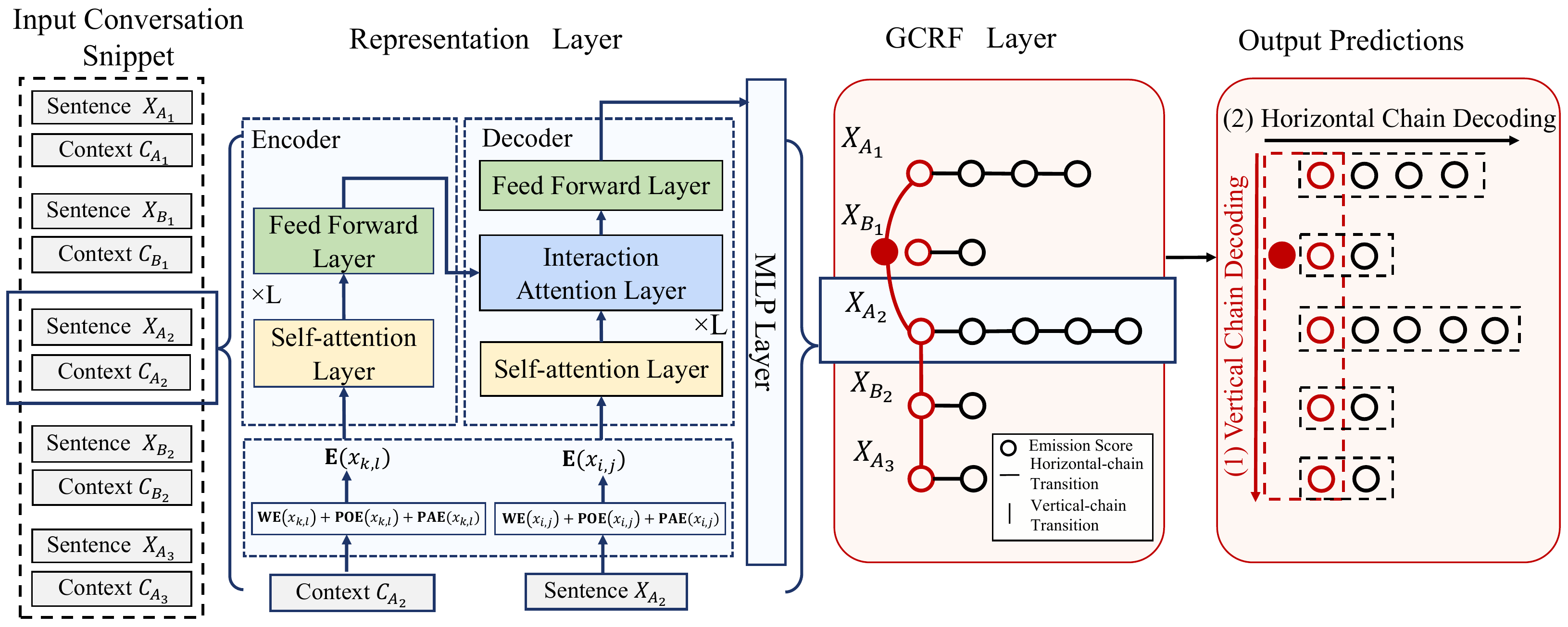}
	\caption{Overall architecture of our Transformer-GCRF model.}
	\label{framework}
\end{figure*}

\section{Related Work}
\subsection{Dropped pronoun recovery}
As pronouns are frequently dropped in informal genres, \citet{Yang:15} first introduced dropped pronoun recovery as an independent task and used a Maximum Entropy classifier to recover DPs in text messages. \citet{giannella:17} employed a linear-chain CRF to jointly predict the position, person, and number of the dropped pronouns in a single utterance, 
to exploit the sequential nature of this problem.
With the powerful representation capability of neural network~\citep{Xu:FNTIR:DeepMatch}, \citet{zhang:neural} introduced a MLP neural network to recover the dropped pronouns based on the concatenation of word embeddings within a fixed-length window. \citet{Yang2019NDPR} proposed a neural network with structured attention to model the interaction between dropped pronouns and their referents using both sentence-level and word-level context, and again each dropped pronoun is predicted independently. 
\citet{tong2019} further incorporated specific external knowledge to identify the referent more accurately. 
None of these methods consider the dependencies between pronouns in neighboring utterances.

\subsection{Zero pronoun resolution}
Zero pronoun resolution~\citep{Zhao:07,Kong:10,Chen:16,Yin:17,yin:2018} is a line of research closely related to dropped pronoun recovery. The difference between these two tasks is that zero pronoun resolution focuses on resolving  anaphoric pronouns to their antecedents assuming the position of the dropped pronoun is already known. However, in dropped pronoun recovery, we consider both anaphoric and non-anaphoric pronouns, and attempt to recover the type of dropped pronoun but not its referent.~\citet{su2019improving} also presented a new utterance rewriting task which improves the multi-turn dialogue modeling through recovering missing information with coreference. 

\subsection{Conditional random fields}
Conditional Random Fields (CRFs) are commonly used in sequence labeling. It models the conditional probability of a label sequence given a corresponding sequence of observations.
\citet{lafferty2001conditional} made a first-order Markov assumption among labels and proposed a linear-chain structure that can be decoded efficiently with the Viterbi algorithm. 
\citet{sutton2004dynamic} introduced dynamic CRFs to model the interactions between two tasks and jointly solve the two tasks when they are conditioned on the same observation. \citet{zhu20052d} introduced two-dimensional CRFs to model the dependency between neighborhoods on a 2D grid to extract object information from the web. \citet{sutton2012introduction} also explored how to generalize linear-chain CRFs to general graphs.
CRFs have also been combined with powerful neural networks to tackle sequence labeling problems in NLP tasks such as POS tagging and Named Entity Recognition (NER)~\cite{lample:16,ma:2016,liu:18empower}, but 
existing research has not explored how to combine deep neural networks with general CRFs.

\section{Our Approach: Transformer-GCRF}
We start by formalizing the dropped pronoun recovery task as follows. Given a Chinese conversation snippet $\mathrm{X}=(\mathbf{x}_1,\cdots, \mathbf{x}_n)$ which consists of $n$ pro-drop utterances, where the $i$-th utterance $\mathbf{x}_i = (x_{i1},\cdots, x_{im_i})$ is a sequence of $m_i$ tokens, and additionally given a set of $k$ possible labels $\mathcal{Y} = \{y_1, \cdots, y_{k-1}\}\cup \{\mathrm{None}\}$ where each $y_j$ corresponds to a pre-defined pronoun~\citep{Yang:15} or `None', which means no pronoun is dropped, the goal of our task is to assign a label $y\in \mathcal{Y}$ to each token in $\mathrm{X}$ to indicate whether a pronoun is dropped before this token and the type of pronoun. We model this task as the problem of maximizing the conditional probability $p(\mathrm{Y|X})$, where $\mathrm{Y}$ is the label sequence assigned to the tokens in $\mathrm{X}$. 
The conditional probability of a label assignment $\mathrm{Y}$ given the whole conversation snippet $\mathrm{X}$ can be written as:
\[
p(\mathrm{Y|X}) = \frac{e^{s(\mathrm{X,Y})}}{\sum_{\widetilde{\mathrm{Y}} \in \mathcal{Y}_\mathrm{X}} e^{s(\mathrm{X,\widetilde{Y}})}},
\]
where $s(\mathrm{X,Y})$ denotes score of the sequences of predictions in the conversation snippet. The denominator is known as partition function, and $\mathcal{Y}_\mathrm{X}$ contains all possible tag sequences for the conversation snippet $\mathrm{X}$.

\subsection{Overview of Transformer-GCRF}
We score each pair of $(\mathrm{X},\mathrm{Y})$ with our proposed Transformer-GCRF, as shown in Figure~\ref{framework}. When pre-processing the inputs, we attach a context to each pro-drop utterance $\mathbf{x}_n$ in the snippet $\mathrm{X}$. The context $C_n = \{\mathbf{x}_{n-5},... \mathbf{x}_{n-1},  \mathbf{x}_{n+1}, \mathbf{x}_{n+2}\}$ consists of the previous five utterances as well as the next two utterances following the practices in~\citep{Yang2019NDPR}, and provides referent related contextual information to help recover the dropped pronouns. 
The representation layer uses the Transformer structure to encode the context $C_n$ and generates representations for tokens in utterance $\mathbf{x}_n$ from the decoder.
The prediction layer then utilizes a generalized CRF to model the cross-utterance and inter-utterance dependencies between the predictions in the conversation snippet, and outputs the predicted sequence for tokens in the snippet.

\subsection{Representation layer}
We employ the encoder-decoder structure of Transformer~\citep{vaswani2017attention} to generate the representations for the tokens in pro-drop utterance $\mathbf{x}_i$ and context $C_i$ separately.

\subsubsection{Context encoder}
\label{sec:encoder}
The context encoder first unfolds all tokens in the context $C_i$ into a linear sequence as: $(x_{i-5,1}, x_{i-5,2}, ..., x_{i+2,m_{i+2}})$, and then inserts the delimiter `[SEP]' between each pair of utterances. Following the Transformer model~\cite{vaswani2017attention}, the input embedding of each token $x_{k,l}$ is the sum of its word embedding $\bm{\mathrm{WE}}(x_{k,l})$, position embedding $\bm{\mathrm{POE}}(x_{k,l})$, and speaker embedding $\bm{\mathrm{PAE}}(x_{k,l})$ as:
\begin{equation}\nonumber
\resizebox{0.89\hsize}{!}{$
\bm{\mathrm{E}}(x_{k,l}) = \bm{\mathrm{WE}}(x_{k,l}) + \bm{\mathrm{POE}}(x_{k,l}) + \bm{\mathrm{PAE}}(x_{k,l})
$}.
\end{equation} 

The token embeddings $\bm{\mathrm{E}}(x_{k,l})$ are then fed into the encoder, which is a stack of $L$ encoding blocks. Each block contains two sub-layers (i.e., a self-attention layer and a feed-forward layer) as:
\begin{equation}\label{eq:ContextRep}
\resizebox{0.89\hsize}{!}{$
\bm{\mathrm{H}}^{(l)} = \mathrm{FNN(SelfATT}(\bm{\mathrm{H}}^{(l-1)}_Q, \bm{\mathrm{H}}^{(l-1)}_K, \bm{\mathrm{H}}^{(l-1)}_V)),
$}
\end{equation}  
for $l = 1, \cdots, L$, where `FNN' and `SelfATT' denotes the feed-forward and self-attention networks respectively, and
\[
\begin{split}
\resizebox{0.89\hsize}{!}{$
\bm{\mathrm{H}}^{(0)} = [\bm{\mathrm{E}}(x_{i-2,1}), \bm{\mathrm{E}}(x_{i-2,2}), \cdots, \bm{\mathrm{E}}(x_{i+1,m_{i+1}})].
$}
\end{split}
\]

In Equation~\ref{eq:ContextRep}, the self-attention layer first projects the input as a query matrix ($\bm{\mathrm{H}}^{(l-1)}_Q$), a key matrix ($\bm{\mathrm{H}}^{(l-1)}_K$), and a value matrix ($\bm{\mathrm{H}}^{(l-1)}_V$). A multi-head attention mechanism is then applied to these three matrices to encode the input tokens in the context.

\subsubsection{Utterance decoder}
To generate the representations for tokens in the pro-drop utterance $\mathbf{x}_i$ and exploit referent information from its context $C_i$, we utilize the decoder component of the Transformer to represent $\mathbf{x}_i$. Similar to the context encoder, the inputs to the utterance decoder are the embeddings of the tokens. Each embedding $\bm{\mathrm{E}}(x_{i,j})$ is also a sum of its word embedding, position embedding, and speaker embedding.
Then, the input to the decoder, denoted as $\bm{\mathrm{S}}^{(0)}$, is a concatenation of all the token embeddings: 
\[
\resizebox{0.89\hsize}{!}{$
\bm{\mathrm{S}}^{(0)}_i = [\bm{\mathrm{E}}(x_{i,1}), \bm{\mathrm{E}}(x_{i,2}), \cdots, \bm{\mathrm{E}}(x_{i,{m_i}})].
$}
\]

The decoder is still a stack of $L$ decoding blocks. Each decoding block $\mathrm{Dec}(\cdot)$ contains three sub-layers (i.e., a self-attention layer, an interaction attention layer, and a feed-forward layer) as:
\[
\begin{split}
\bm{\mathrm{S}}^{(l)}_i = &\mathrm{Dec}(\bm{\mathrm{S}}^{(l-1)}_i, \mathbf{H}^{(L)}_i)\\
= & \mathrm{FFN}(\mathrm{InterATT}(\mathrm{SelfATT}(\bm{\mathrm{S}}^{(l-1)}_i), \mathbf{H}^{(L)}_i)),
\end{split}
\]
for $l = 1,\cdots, L$, where $\mathrm{FFN}$ is a feed-forward network, $\mathrm{SelfATT}$ is a self-attention network.

\begin{figure*}
	\centering
	\includegraphics[width=13.7cm, height=3.8cm]{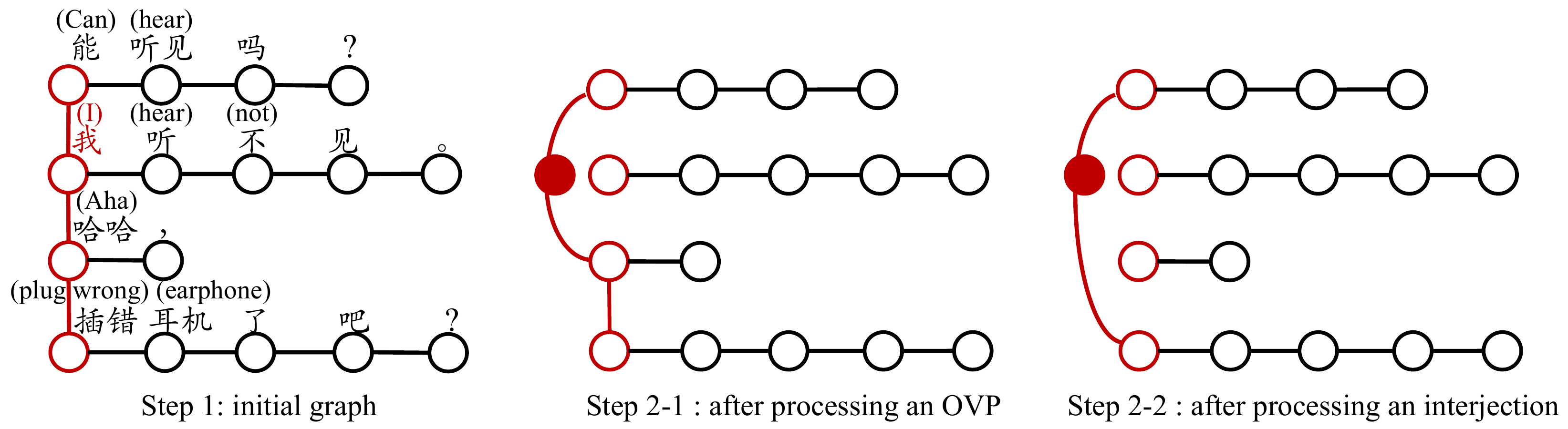}
	\caption{The GCRF graph construction. Step 1 constructs a initial graph. The tokens in each utterance are shown and the nodes corresponding to the first token in each utterance are highlighted in red; step 2-1 processes an OVP (in the second utterance) and adds an observed (shaded) node for token ``我/(I)''; step 2-2 processes an interjection (in the third utterance) and skips the node corresponding to the token ``哈哈/(Aha)''.}
	\label{graph}
\end{figure*}

Finally, the output states of the decoder $\bm{\mathrm{S}}^{(L)}$ are transformed into logits through a two-layer MLP network as:
\begin{equation}
\bm{\mathrm{P}} = \bm{\mathrm{W}}_1 \cdot \tanh(\bm{\mathrm{W}}_2 \cdot \bm{\mathrm{S}}^{(L)}+b_2)+b_1,
\end{equation}
where the logits matrix $\bm{\mathrm{P}}$ of size $n\times m\times k$ will be fed into a subsequent prediction layer. $k$ is the number of distinct tags, and each element $\mathrm{P}_{i,j,l}$ refers to the emission score of the $l$-th tag of the $j$-th word in the $i$-th utterance.

\subsection{GCRF layer}
We utilize an elaborately designed general conditional random fields (GCRF) layer to recover dropped pronouns by modeling cross-utterance and intra-utterance  dependencies between dropped pronouns.

\subsubsection{Graph construction in GCRF}\label{sec:GC}
Given a conversation snippet, a graph is constructed where
each node, corresponding to a token, is a random variable $y$ that represents the type of the pronoun defined in $\mathcal{Y}$. 
The edges in the graph are defined by the following two steps: 

\noindent \textbf{Step 1: Initial graph construction:}  
We first split each compound utterance into several simple utterances by punctuation, and connect the nodes corresponding to the tokens in the same simple utterance with horizontal edges to model intra-utterance dependencies. Then we link the first tokens in consecutive utterances with a vertical chain to model the cross-utterance  dependencies.
Step 1 in Figure~\ref{graph}  shows an initial graph for a conversation snippet.

\noindent \textbf{Step 2: Vertical edge refinement:} Though the vertical chain constructed in Step 1 can capture most of the cross-utterance dependencies, they can be further refined considering the following two general cases in conversation:
\begin{itemize}

\setlength{\itemsep}{-2pt}%
\item \textbf{Overt pronouns (OVP):} If an OVP appears as the first token in a utterance, it is clear that there is a dependency between the OVP and the dropped pronoun in neighboring utterances. To model this phenomenon, an observed node (with the value of its pronoun type) is inserted in the graph, and the vertical chain linked to the original node is moved to this new node. Step 2-1 in Figure~\ref{graph}  shows the refined graph after OVPs are processed. 
\item \textbf{Interjections:} If the first token in an utterance is an interjection (e.g., ``嗯 / Well'', ``哈哈 / Aha'' etc.), it is better to skip the utterance in the vertical chain because the short utterance consisting of only interjections and punctuation does not provide useful information about the dependencies between pronouns.
Step 2-2 in Figure~\ref{graph}  shows the refined graph after interjections are processed. 
\end{itemize}

\subsubsection{Pronoun prediction}
It is obvious that the GCRF is a special case of the 2D CRFs. To predict the labels of the nodes following the practices in~\citep{zhu20052d}, we employ a modified Viterbi algorithm in which the nodes in the vertical chain are decoded first. Specifically, the constructed graph consists of two types of cliques: one from the horizontal chains and the other from the vertical chain. 
Given the emission score matrix $\bm{\mathrm{P}}$ outputted from the decoder layer (see Section~\ref{sec:encoder}), the joint score $s(\mathrm{X,Y})$ of the predictions 
can be computed by first computing the sum of horizontal chains and then summing up scores of the transitions in the vertical chain as:
\begin{equation}
\label{score}
    s(\mathrm{X,Y})=\sum\limits_{i=1}^{n}s_{hi} + \sum\limits_{i=1}^{n-1}\mathrm{A}_{T_i, T_{i+1}}^{(2)},
\end{equation}
\begin{equation}\nonumber
\label{score_h}
    s_{hi}=\sum\limits_{j=1}^{m-1}\mathrm{A}_{y_{i,j}, y_{i,j+1}}^{(1)}+\sum\limits_{j=1}^{m}\mathrm{P}_{i,j,y_{i,j}},
\end{equation}
where $\bm{\mathrm{A}}^{(1)}$ and $\bm{\mathrm{A}}^{(2)}$ are the transition matrices of the horizontal chains and the vertical chain, respectively;  $\mathrm{A}_{i,j}$ indicates the transition score from tag $i$ to tag $j$; and the node $T_i$ is defined as, 
\begin{equation}\nonumber
\label{case}
T_i=\begin{cases}
  y_{\mathrm{OVP}}&\textrm{if the node is an observed OVP}\\
  y_{i,1}& \textrm{otherwise}\\
\end{cases}
\end{equation}
where $y_{\mathrm{OVP}}\in \mathcal{Y}$ is the observed label corresponds to the specific OVP. 
The first term in Eq.~(\ref{score}) is the score corresponding to the horizontal chain cliques, and the second term corresponds to the vertical chain clique.

\subsection{Decoding the GCRF and Model training}
The sequence that maximizes the conditional probability $p(\mathrm{{Y}|X})$ is outputted as the prediction: 
\begin{equation}\nonumber
\mathrm{Y}^* =\mathop{\arg\max}_{{\mathrm{Y}} \in \mathcal{Y}_\mathrm{X}} p(\mathrm{{Y}|X}).
\end{equation}

A modified Viterbi algorithm is used to find the best labeling sequence. Specifically, we first applies the Viterbi algorithm to decode the vertical chain. Then, the vertical chain decoding results are used as the observed nodes in the graph, and the standard Viterbi algorithm is applied to each horizontal chain in parallel. Algorithm~\ref{alg:Framework} shows the Transformer-GCRF decoding process.

Given a set of labeled conversation snippets $\mathcal{D}$, the model parameters are learned by jointly maximizing the overall log-probabilities of the ground-truth label sequences:

\vspace{6pt}
\hspace{30pt} $\max \sum_{\mathrm{(X,Y)}\in \mathcal{D}} \log(p{\rm (Y|X)}).$
\vspace{6pt}

\begin{algorithm}[t!]
\caption{Transformer-GCRF Decoding.}
\label{alg:Framework}
\begin{algorithmic}[1]
 \Require The emission score matrix $\bm{\mathrm{P}}$; Transition matrices $\bm{\mathrm{A}}^{(1)}$ and $\bm{\mathrm{A}}^{(2)}$.
 \Ensure The best path $\mathrm{Y}^*$
 \For{$i=1,\dots,n$}
 \State $s_{hi}$, $\bm{bp}_i$ $\leftarrow$ ForwardScore($\bm{\mathrm{P}}_i$, $\bm{\mathrm{A}}^{(1)}$)
 \EndFor 
 \State $\bm{\mathrm{P}}_h = [s_{h1}, s_{h2}, \cdots, s_{hn}]$
 \State $s(\mathrm{X,Y})$, $\bm{bp}_v$ $\leftarrow$ ForwardScore($\bm{\mathrm{P}}_h$, $\bm{\mathrm{A}}^{(2)}$)
 \State $\mathrm{Y}^*_{n,1}$ $\leftarrow$ $\mathop{\arg\max}{(s(\mathrm{X,Y}))}$
 \State \{$\mathrm{Y}^*_{1,1},\cdots, \mathrm{Y}^*_{n-1,1}$\}$\leftarrow$TraceBack ($\mathrm{Y}^*_{n,1}$, $\bm{bp}_v$)
 \For{$i=1,...,n$}
 \State\{$\mathrm{Y}^*_{i,2},\cdots,\mathrm{Y}^*_{i,m_i}$\}$\leftarrow$TraceBack($\mathrm{Y}^*_{i,1}$, $\bm{bp}_i$)
 \EndFor 
 \Function{TraceBack}{$y$, $\bm{bp}$}
 \State $t = length(\bm{bp})$
 \State $z_1$ $\leftarrow$ $y$
 \For{$j=2, \cdots, t$}
 \State $z_j$ $\leftarrow$ $bp_{j, z_{j-1}}$
 \EndFor
 \State \Return \{$z_2, \cdots, z_t$\}
 \EndFunction
 \State \Return $\mathrm{Y}^*$
 \end{algorithmic}
\end{algorithm}

\section{Datasets and Experimental Setup}
\textbf{Datasets:} \quad We evaluate the performance of Transformer-GCRF on three conversation benchmarks: Chinese text message dataset (SMS), OntoNotes Release 5.0, and BaiduZhidao. The SMS dataset is described in~\cite{Yang:15} and contains 684 text message documents generated by users via SMS or Chat. Following~\cite{Yang:15,Yang2019NDPR}, we reserved 16.7\% of the training set as the development set, and a separate test set was used to evaluate the models. The OntoNotes Release 5.0 was released in the CoNLL 2012 Shared Task. We used the TC section which consists of transcripts of Chinese telephone conversation speech. The BaiduZhidao dataset is a question answering dialogue corpus collected by~\cite{zhang:neural}. Ten types of  dropped pronouns are annotated according to the pronoun annotation guidelines.
The statistics of these three benchmarks are reported in Table~\ref{dataset statistic}.

\begin{table}
\begin{center}
\begin{tabular}{p{0.9cm}<{\centering}|p{1.45cm}<{\centering}c|p{1.45cm}<{\centering}c}
\Xhline{1.2pt}
& \multicolumn{2}{c|}{Training} & \multicolumn{2}{c}{Test} \\ \cline{2-5}
& \#Sentences & \#DPs & \#Sentences & \#DPs \\ \hline
SMS & 35,933 & 28,052 & 4,346 & 3,539\\
TC & 6,734 & 5,090 & 1,122 & 774\\
Zhidao & 7,970 & 5,097 & 1,406 & 786\\
\Xhline{1.2pt}
\end{tabular}
\end{center}
\caption{\label{dataset statistic} Statistics of training and test sets on three conversational benchmarks.}
\end{table}

\begin{table*}[t!]
 	\begin{center}
 		\begin{tabular}{p{4.8cm}|p{0.7cm}<{\centering} p{0.7cm}<{\centering} p{0.9cm}<{\centering}| p{0.7cm}<{\centering} p{0.7cm}<{\centering} p{0.9cm}<{\centering}|p{0.7cm}<{\centering} p{0.7cm}<{\centering} p{0.9cm}<{\centering}}
 			\Xhline{1.2pt}
 			\multirow{2}{*}{Model} & \multicolumn{3}{c|}{Chinese SMS} & \multicolumn{3}{c|}{TC of OntoNotes} & \multicolumn{3}{c}{BaiduZhidao} \\ \cline{2-10} 
			 &  P(\%) &  R(\%) &  F & P(\%) & R(\%) & F & P(\%) & R(\%) & F \\ \hline
			MEPR~\cite{Yang:15} & 37.27 & 45.57 & 38.76 & - & - & - & - & - & - \\ 
			NRM~\cite{zhang:neural} & 37.11 & 44.07 & 39.03 & 23.12 & 26.09 & 22.80 & 26.87 & 49.44 & 34.54 \\ 
			BiGRU & 40.18 & 45.32 & 42.67 & 25.64 & 36.82 & 30.93 & 29.35 & 42.38 & 35.83 \\ 
			NDPR~\cite{Yang2019NDPR} & 49.39 & 44.89 & 46.39 & 39.63 & 43.09 & 39.77 & 41.04 & 46.55 & 42.29 \\ \hline
			NDPR-GCRF & 51.27 & 45.45 & 47.73 & 39.45 & 43.55 & 40.53 & 39.60 &\bf 49.54 & 43.39 \\ 
			Transformer & 51.53 & 46.18 & 48.21 & 39.91 & 43.98 & 41.79 & 42.13 & 46.63 & 43.58 \\
			Transformer-GCRF(w/o refine) & 52.32 & 47.50 & 49.23 & 40.18 & 44.02 & 42.01 & 42.41 & 47.76 & 43.62 \\ \hline
			Transformer-GCRF &\bf 52.51 &\bf 48.12 &\bf 49.81$^{*}$ &\bf 40.48 &\bf 44.64 &\bf 42.45$^{*}$ &\bf 43.30 & 46.54 &\bf 43.92$^{*}$ \\ \Xhline{1.2pt}
		\end{tabular}
 	\end{center}
 	\caption{\label{comparable-result} Results in terms of precision, recall and F-score produced by the baseline systems and variants of our proposed Transformer-GCRF framework. `$*$' indicates the improvement over the best baseline NDPR is significant (t-tests and $p$-value $\leq 0.05$).}
\end{table*}

\textbf{Baselines:} \quad State-of-the-art dropped pronoun recovery models are used as baselines: (1) MEPR~\cite{Yang:15} which leverages a set of elaborately designed features  and trains a Maximum Entropy classifier to predict the type of dropped pronoun before each token; 
(2) NRM~\cite{zhang:neural} which employs two separate MLPs to predict the position and type of a dropped pronoun utilizing representation of words in a fixed-length window;
(3) BiGRU which utilizes a bidirectional RNN to encode each token in a pro-drop sentence and makes prediction based on the encoded states;
(4) NDPR~\cite{Yang2019NDPR} which models dropped pronoun referents by attending to the context and independently predicts the presence and type of DP for each token.

We also compare three variants of Transformer-GCRF as: 
(1) Transformer-GCRF(w/o refine) which removes Step 2 in Section~\ref{sec:GC} during the graph construction process, for exploring the effectiveness of processing OVP and interjections;
(2) Transformer which removes the whole GCRF layer that globally optimizes the prediction sequences, and directly adds a MLP layer on the top of Transformer encoder to predict the dropped pronouns. It aims to explore the contribution of Transformer encoder among the total effectiveness of Transformer-GCRF;
(3) NDPR-GCRF which replaces the Transformer structure in the presentation layer with the NDPR model~\cite{Yang2019NDPR}.

\textbf{Training details:} \quad In all of the experiments, a vocabulary was first generated based on the entire dataset, and the out-of-vocabulary words are represented as ``UNK''. 
The length of utterances in a conversation snippet is set as 8 in our work.
In Transformer-GCRF, both the encoder and decoder in the Transformer have 512 units in each hidden layer. 
We augment each utterance with a context consisting of seven neighboring utterances according to the practice in~\citep{Yang2019NDPR}.
In each experiment, we trained the model for 30 epochs on one GPU, which took more than five hours, and the model with the highest F-score on the development set was selected for testing. 
Following ~\cite{glorot2010understanding}, in all of the experiments the weight matrices were initialized with uniform samples from $[-\sqrt{\frac{6}{r+c}}, +\sqrt{\frac{6}{r+c}}]$, where $r$ and $c$ are the number of rows and columns in the corresponding matrix.
Adam optimizer \cite{kingma2015adam:} is utilized to conduct the optimization.

\section{Results and Analysis}
\subsection{Performance Evaluation}
We apply our Transformer-GCRF model to all three conversation datasets to demonstrate the effectiveness of the model. Table~\ref{comparable-result} reports the results of our Transformer-GCRF model as well as the baseline models in terms of precision (P), recall (R), and F-score (F). 

From the results, we can see that our proposed model and its variants outperformed the baselines on all datasets. The best model Transformer-GCRF achieves a gain of 2.58\% average absolute improvement across all three datasets in terms of F-score. We also conducted significance tests on all three datasets in terms of F-score. The results show that our method significantly outperforms the best baseline NDPR ($p<0.05$). 
The proposed Transformer-GCRF suffers from performance degradation when Step 2 is removed from the graph construction process (i.e., referring to the results of Transformer-GCRF(w/o refine) in Table~\ref{comparable-result}),
which demonstrates the important role of OVPs in modeling dependencies between different utterances, and the contribution of noise reduction resulting from skipping short utterances starting with interjections. 
Both our proposed Transformer-GCRF model and the variant Transformer-GCRF(w/o refine) model outperform the variant Transformer, which demonstrates that the effectiveness comes from not only the powerful Transformer encoder, but also the elaborately designed GCRF layer. Moreover, the variant NDPR-GCRF, which encodes the pro-drop utterances with BiGRU as NDPR~\citep{Yang2019NDPR}, still outperforms the original NDPR. This shows that the proposed GCRF is effective in modeling cross-utterance dependencies regardless of the underlying representation.

\begin{figure}[t!]
	\centering
	\includegraphics[width=7.7cm, height=6.3cm]{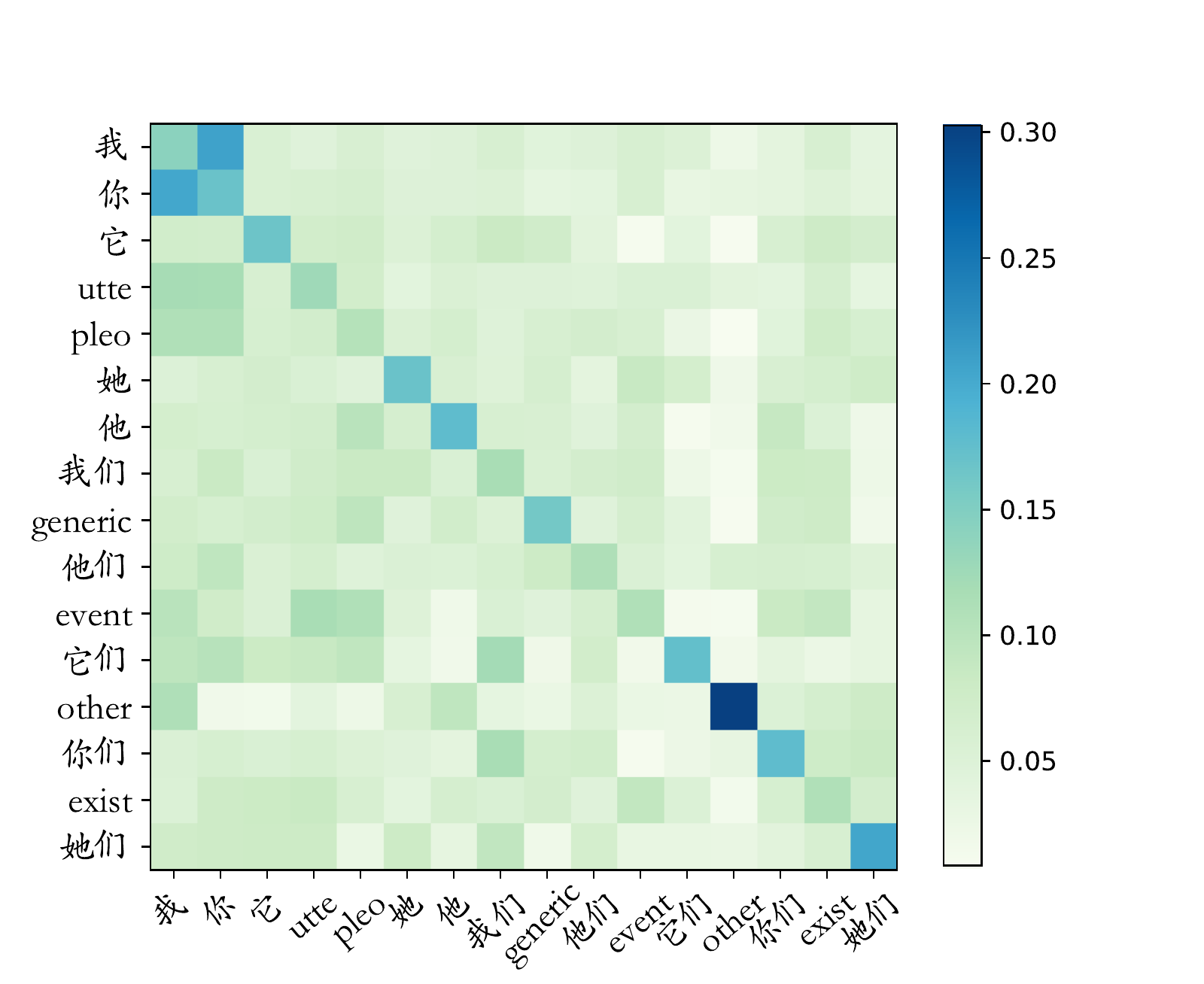}
	\caption{Visualization of the transition weight between each pair of pronouns among 16 types of pre-defined pronouns (i.e., except the category `None'), obtained from the vertical chain transition matrix $\bm{\mathrm{A}}^{(2)}$. Darker color indicates higher transition weight between these two types of pronouns.}
	\label{transition}
\end{figure}

\subsection{Motivation and Effects of GCRF}
\label{demonstrate:GCRF}
\subsubsection{Motivation by statistical results}
The GCRF model is motivated with a quantitative analysis of our data, which shows that 79.6\% of the dropped pronouns serve as the subject of a sentence, and occur at utterance-initial positions. The pronouns dropped at the beginning of consecutive utterances are strongly correlated with  dialogue patterns and thus modeling conversational structures helps improve recover dropped pronouns.
Other pronouns dropped as objects in the middle of a utterance should be recovered by modeling intra-utterance dependencies.

To further explore the cross-utterance pronoun dependencies,
we collected all 
pronoun pairs occurring at the beginning of consecutive utterances and classified the dependencies 
into one of the three dialogue transitions defined in \cite{xue2016annotating}. We found that 27.33\% of the pairs correspond to \textit{reply} transition, where the second utterance is a response to the first utterance, and 18.60\% of pairs correspond to the \textit{acknowledgment} transition, where the second utterance is an acknowledgment of the first utterance. In both cases, the utterances involve a shift of speaker, which is accompanied by a shift in the use of personal pronouns. Another 47.79\% of the pairs correspond to the \textit{expansion} transition, where the second utterance is an elaboration of the first utterance and the same pronoun is used. 

\subsubsection{Visualizing transition matrix of GCRF}
To investigate whether our GCRF model actually learned the dependencies revealed by the quantitative analysis of our corpus, 
we visualize the transition matrix $\bm{\mathrm{A}}^{(2)}$ of the vertical chain in Figure~\ref{transition}. We can see that the learned transition matrix matches well with the distribution of dialogue patterns. The matrix shows that the higher transition weights on diagonal correspond to the strong \textit{expansion} transition in which the same pronoun is used in consecutive utterances and the transition weights between ``我 (I)'' and ``你 (you)'' (top-left corner) are high as well, indicating the strong \textit{reply} transition. Moreover, the \textit{acknowledgement} transition usually exists from the pronoun ``previous utterance'' to ``我 (I)'' or ``你 (you)''.

\begin{figure}[t!]
	\centering
	\includegraphics[width=6.8cm, height=5.4cm]{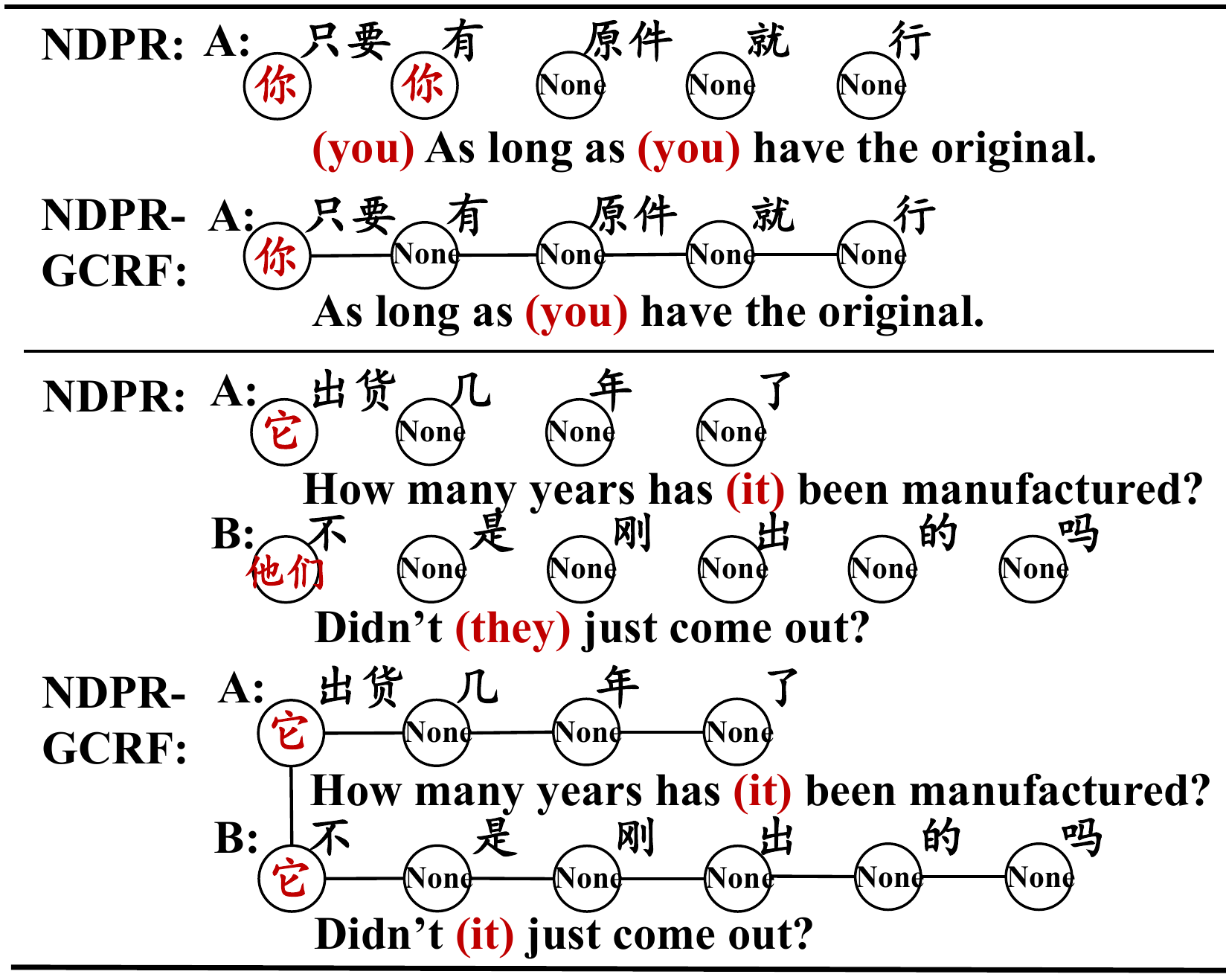}
	\caption{Example results of NDPR and NDPR-GCRF. The recovered pronouns are marked with red color and shown in brackets. 
	}
	\label{case_study}
\end{figure}

\subsubsection{Case studies}
We demonstrate the effectiveness of GCRF by
comparing the outputs of NDPR and NDPR-GCRF on the entire test set, and present some concrete cases in Figure~\ref{case_study}. The examples show that the horizontal chains in GCRF contributes by preventing redundant predictions in the same utterance. For example, in the first case, the second pronoun ``你 (you)'' is repeatedly recovered by NDPR since the dependency between the predictions of the first two tokens is ignored.
The vertical chain contributes by predicting coherent dropped pronouns at the beginning of the utterances. For example, in the second case, the second utterance is a \textit{reply} of the first one, and NDPR-GCRF recovers these two pronouns correctly by considering their dependency.

\subsection{Effects of the Transformer architecture}
\begin{figure}[t!]
	\centering
	\includegraphics[width=7.5cm, height=5.0cm]{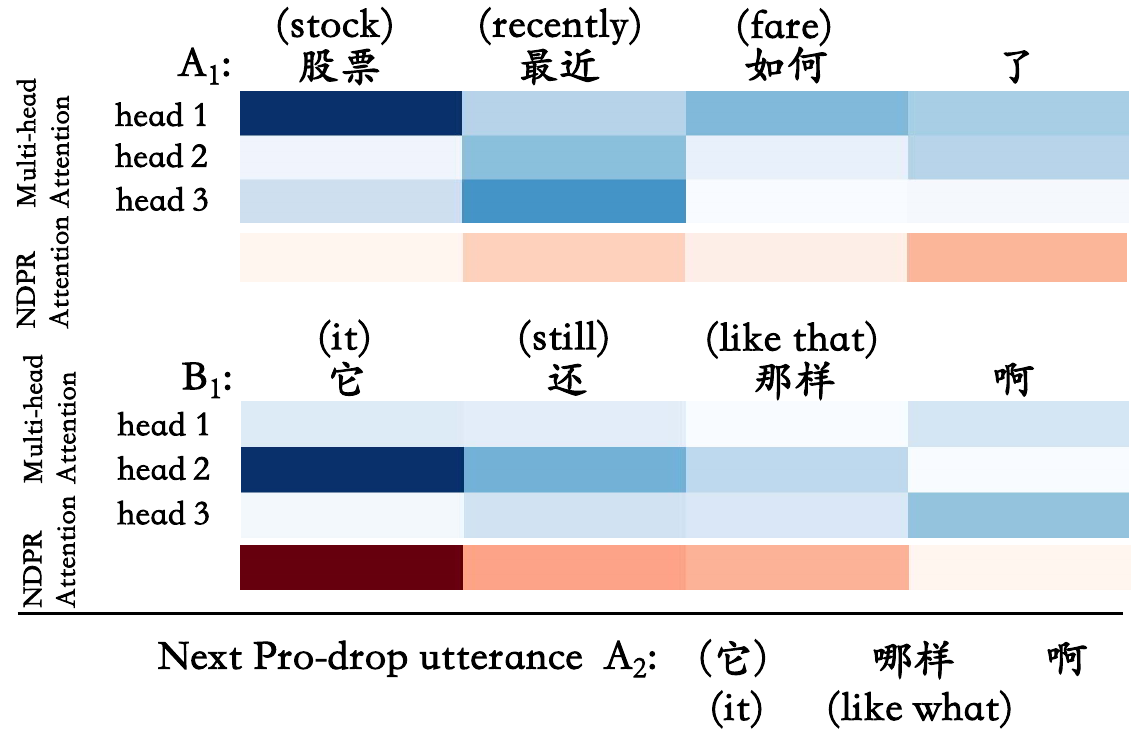}
	\caption{Visualization of multi-head attention in Transformer-GCRF and structured attention in NDPR.}
	\label{attention}
\end{figure}

We further study the effectiveness of multi-head attention in Transformer structure.
Figure~\ref{attention} shows an example conversation snippet with three utterances and the pronoun ``它 (it)'' in the last utterance is dropped. The Transformer's attention weights corresponding to three heads which are shown in blue, and the NDPR's attention weights are shown in brown. From the results, we can see that ``head 1'' is responsible for associating ``股票 (stock)'' with ``它 (it)'' (in utterance $A_1$), ``head 2'' is responsible for associating ``它 (it)'' with ``它 (it)'', and ``head 3'' is responsible for collecting noisy information, which is helpful for the training process~\citep{michel2019are,correia2019adaptively}. This is consistent with the observation in~\citep{vig2019a} that multi-head attention is powerful because it uses different heads to capture different relations. NDPR, on the other hand, captures all these the relations with a single attention structure. The results explain why Transformer is suitable for dropped pronoun recovery.

\subsection{Error Analysis}
\begin{figure}[t!]
	\centering
	\includegraphics[width=7.5cm, height=4.7cm]{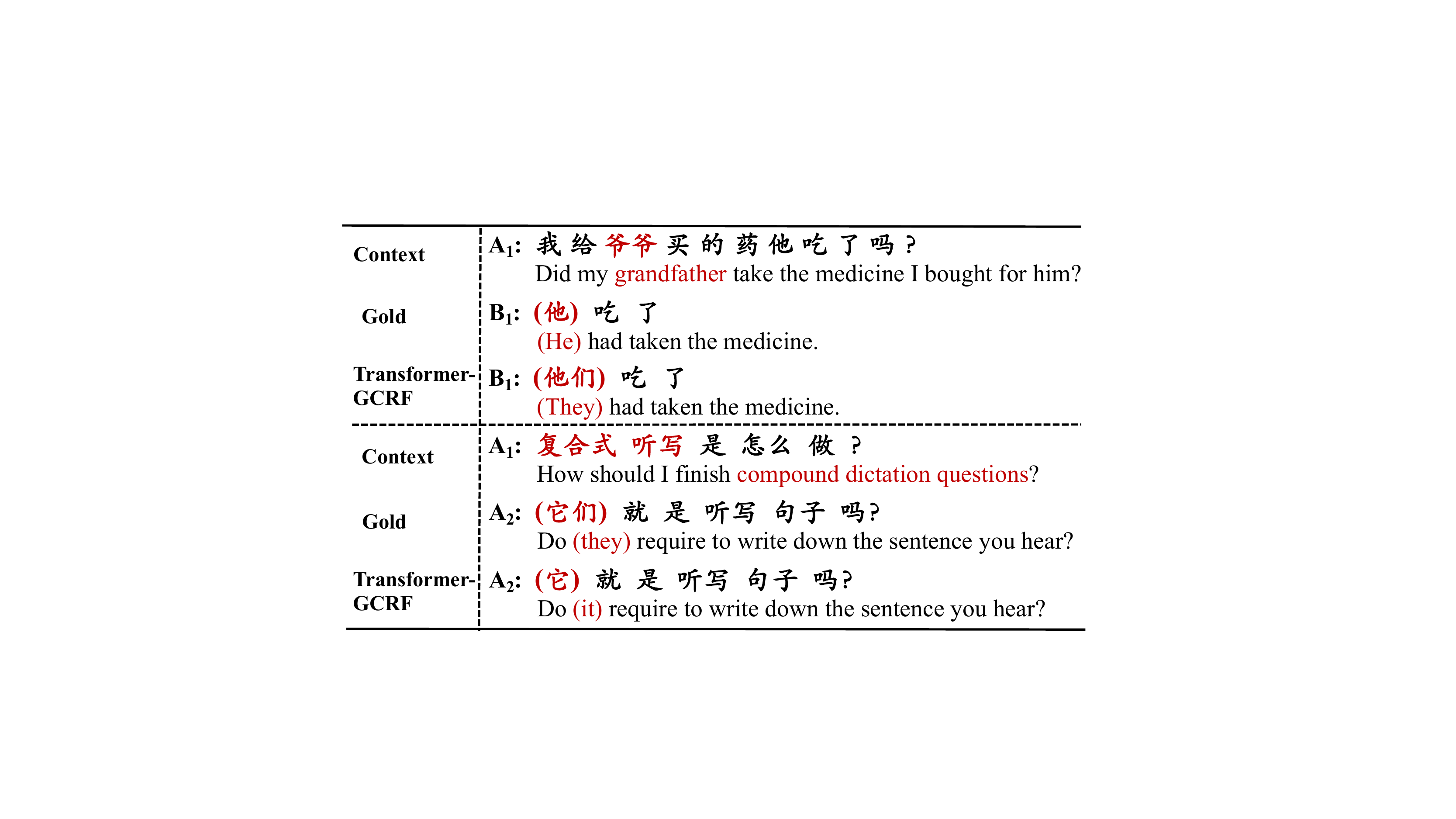}
	\caption{Example errors made by Transformer-GCRF.}
	\label{error_case}
\end{figure}

Besides conducting the performance evaluation and analyzing the effects of different components, we also investigate some typical mistakes made by our Transformer-GCRF model. 
The task of recovering dropped pronouns consists of first identifying the referent of each dropped pronoun from the context and then recovering the referent as a concrete Chinese pronoun based on the referent semantics. Existing work has focused on modeling referent semantics of the dropped pronoun from context, and globally optimizing the prediction sequences by exploring label dependencies. However, there is also something need to do about how to recover the referent as a proper pronoun based on the referent semantics. For example, in two cases of Figure~\ref{error_case}, the referents of the dropped pronouns are correctly identified, while the final pronoun was recovered as ``(他们/they)'' and ``(它/it)'' by mistake. We attribute this to that the model needs to be augmented with some common knowledge about how to recover a referent to the proper Chinese pronoun.

\section{Conclusion and Future Work}
In this paper, we presented a novel model for recovering the dropped pronouns in Chinese conversations. The model, referred to as Transformer-GCRF, formulates  dropped pronoun recovery as a sequence labeling problem. Transformer is employed to represent the utterances and GCRF is used  to make the final predictions, through capturing both cross-utterance and intra-utterance dependencies between pronouns. Experimental results  on three Chinese conversational datasets show that Transformer-GCRF consistently outperforms state-of-the-art baselines. 

In the future, we will do some extrinsic evaluation by applying our proposed model in some downstream applications like pronoun resolution, to further explore the effectiveness of modeling cross-utterance dependencies in practical applications. 

\section{Acknowledgments}
This work was supported by the National Natural Science Foundation of China (No.61702047, No.61872338, No.61832017, No.62006234), Beijing Academy of Artificial Intelligence (No.BAAI2019ZD0305), Beijing Outstanding Young Scientist Program (No.BJJWZYJH012019100020098) and BUPT Excellent Ph.D. Students Foundation (No.CX2020305).

\bibliography{emnlp2020}
\bibliographystyle{acl_natbib}

\end{CJK*}

\end{document}